# Dynamically enhanced static handwriting representation for Parkinson's disease detection

Moises Diaz, Miguel Angel Ferrer, Donato Impedovo, Giuseppe Pirlo, Gennaro Vessio

**Abstract**

Computer aided diagnosis systems can provide non-invasive, low-cost tools to support clinicians. These systems have the potential to assist the diagnosis and monitoring of neurodegenerative disorders, in particular Parkinson's disease (PD). Handwriting plays a special role in the context of PD assessment. In this paper, the discriminating power of "dynamically enhanced" static images of handwriting is investigated. The enhanced images are synthetically generated by exploiting simultaneously the static and dynamic properties of handwriting. Specifically, we propose a static representation that embeds dynamic information based on: (*i*) drawing the points of the samples, instead of linking them, so as to retain temporal/velocity information; and (*ii*) adding pen-ups for the same purpose. To evaluate the effectiveness of the new handwriting representation, a fair comparison between this approach and state-of-the-art methods based on static and dynamic handwriting is conducted on the same dataset, i.e. PaHaW. The classification workflow employs transfer learning to extract meaningful features from multiple representations of the input data. An ensemble of different classifiers is used to achieve the final predictions. Dynamically enhanced static handwriting is able to outperform the results obtained by using static and dynamic handwriting separately.

## Introduction

Parkinson's disease (PD) is one of the most common neurodegenerative disorders and is a growing health problem. It is mainly characterized by motor symptoms, including akinesia, bradykinesia, rigidity and tremor [15]. Currently, there is no cure, and the gradual decline of the patient can only be managed during the disease progression. However, an early diagnosis of PD could be crucial for the prospect of medical treatment and for evaluating the effectiveness of new drug treatments at prodromal stages.

To this end, a growing interest has developed in computer aided diagnosis. Intelligent systems, in fact, have the potential to assist clinicians at the point of care, thus providing better diagnostic tools, while reducing expenditure on public health. In particular, as writing difficulties in PD patients have been documented for some time, a special role for handwriting in the context of PD assessment can safely be assumed.

Handwriting is a complex activity entailing cognitive, kinesthetic and perceptual-motor components [24], the changes in which can be a promising *biomarker* for the evaluation of PD [5,13]. Indeed, there is evidence to suggest that the automatic discrimination between unhealthy and healthy people can be accomplished on the basis of several features obtained through simple and easy-to-perform handwriting tasks [20]. Developing a handwriting-based decision support system is desirable, as it can provide a complementary, non-invasive, and very low-cost approach to the standard evaluations carried out by clinical experts.

## 1.1. Motivations

Currently, the most popular approach to studying the potentialities of automatic handwriting analysis for PD diagnosis involves the use of dynamic aspects of the handwriting process. Several dynamic features have been used, ranging from traditional kinematic and spatio-temporal variables of handwriting to less common measures based on entropy and signal-to-noise ratios [7,21]. Dynamic handwriting analysis benefits from the use of digitizing tablets and electronic pens. By using these devices, it is straightforward to measure the temporal and spatial variables of handwriting, the pressure exerted over the writing surface, the pen inclination, and the movement of the pen while not in contact with the surface, i.e. in-air.

Since dynamic analysis takes into account not only the geometry of the handwritten pattern, but also the underlying generation process, this approach is usually preferred to static analysis, which relies entirely on recorded images of handwriting. One way of obtaining a static version consists in linking the points of the on-surface trajectory as sampled by the acquisition device. Encouraging results for PD diagnosis using this approach have been recently reported [16].

However, according to others [6,10], better insights can be obtained by using "dynamically enhanced" static handwriting. Instead of simply using the image of the handwritten pattern, less realistic but more discriminating images can be obtained by including additional dynamic information during the generation process.

## 1.2. Contribution

In the light of the observations above, this paper proposes to jointly exploit static and dynamic properties of handwriting to improve PD detection. In particular, we propose a static representation that embeds dynamic information by: drawing only the points of the samples, instead of linking them, to retain temporal/velocity information; and by adding pen-ups for the same purpose. The expectation is that applying dynamically enhanced static handwriting to the problem of discriminating PD from healthy controls may improve the performance obtained from using only static or dynamic features of handwriting.

To evaluate the discriminating power of dynamically enhanced static handwriting, we employed transfer learning to extract meaningful features from the raw data for different handwriting tasks. To this end, a freely available dataset, i.e. PaHaW, is used, as it includes several tasks performed by the same subjects.

These features are then combined into an ensemble of classifiers, each built on top of the feature space of every task. For a fair evaluation, the results obtained with this approach are compared to those obtained by using only the static or dynamic features from the same dataset.

The rest of this paper is structured as follows. Section 2 de- scribes the data used for this study. Section 3 presents the proposed method. Section 4 focuses on the experimental results obtained. Section 5 concludes the work.

## 2. Dataset

The "Parkinson's disease handwriting database" (PaHaW) comprises the data of 37 PD patients and 38 age and gender-matched healthy control (HC) subjects, enrolled at the First Department of Neurology, Masaryk University or at the St. Anne's University Hospital, in Brno, Czech Republic. More details about participants' recruitment and clinical evaluation are available [8].

Participants were requested to complete eight handwriting tasks in accordance with a pre-filled template:

1.Drawing an Archimedes spiral.
2.Writing in cursive the letter l.
3.The bigram le.
4.The trigram les.

5. Writing in cursive the word lektorka ("female teacher" in Czech).
6. porovnat ("to compare").
7. nepopadnout ("to not catch").
8. Writing in cursive the sentence Tramvaj dnes už nepojede ("The tram won't go today").

Since not all subjects performed every task, as in [16] we considered only those participants which succeeded in completing all tasks, i.e. 36 PD and 36 HC.

The signals were recorded using the Wacom Intuos 4M digitizing tablet, overlaid with a blank sheet of paper. The sampling rate was 200 samples per second. The raw data captured by the device were the $x$- and $y$-coordinates of the pen position and their corresponding time coordinates. Moreover, measures of pen inclination, i.e. azimuth and altitude, and the pressure exerted over the writing surface were recorded. The so-called "button status", which has a binary value of 0 for pen-ups (in-air movement) and 1 for pen-downs (on-surface movement), were also noted.

## 3. Proposed method

An overall scheme of the proposed method is depicted in Fig. 1. From the time series signals of each task, (raw) dynamically enhanced static images are generated: this is the main contribution of the present paper. To enrich feature learning, additional representations of the input data (filtered and edge images) are used. Three convolutional neural networks (CNNs), sharing the same architecture, are fed with the generated images so as to learn meaningful features. The CNN features are combined into a high dimensional feature vector, one for each task, and used as input to a traditional machine learning algorithm. The predictions from the various tasks are then pooled in a majority voting scheme to achieve the final classification.

It is worth remarking that this experimental set up is similar to the one proposed for use [16] on the same dataset. The main difference concerns the use of additional dynamic information while, in that earlier work, features were learned from only on-surface movement.

### 3.1. Generation of enhanced images

PaHaW does not include images, but separate files, one for each task, which store the dynamic information. An image of the performed task, however, can be reconstructed by using the $(x, y)$-coordinates of the pen position. As in [16], the $x$-coordinate has been normalized by subtracting the minimum value from every $x$ value recorded. Also, the $y$-coordinate has been normalized by subtracting from each co-ordinate the mean value. In this way, the points can be plotted on a coordinate system and the image can be extracted. The pen position both on the pad surface and in-air has been considered. It is worth noting that on-surface movement was represented in black, and in-air by gray, thus generating a gray-scale image (size $432 \times 288$ pixels) for each task. The choice of using different colors is motivated by the fact that the two hand-writing trajectories (on-surface/in-air) have different meanings: since the convolution filters of the neural network work on different channels, such a difference can be captured by the network.

Note that, contrary to Moetesum et al. [16], the image re-construction has not been made by linearly connecting the $(x, y)$-coordinates of the pen position, but by plotting them. Because of the acquisition sampling frequency, when the speed of writing is low, points are close to each other and strokes seem darker, while, when the writing speed is higher, points are more distant, and strokes seem lighter. This provided us with some information about velocity, thus injecting additional dynamic information during the image reconstruction. An example of an enhanced image is shown in Fig. 2.

As in [10], the synthetic images could have been improved by exploiting other dynamic information, e.g. pressure. How to further improve the generation of synthetic handwriting for PD classification calls for future research.

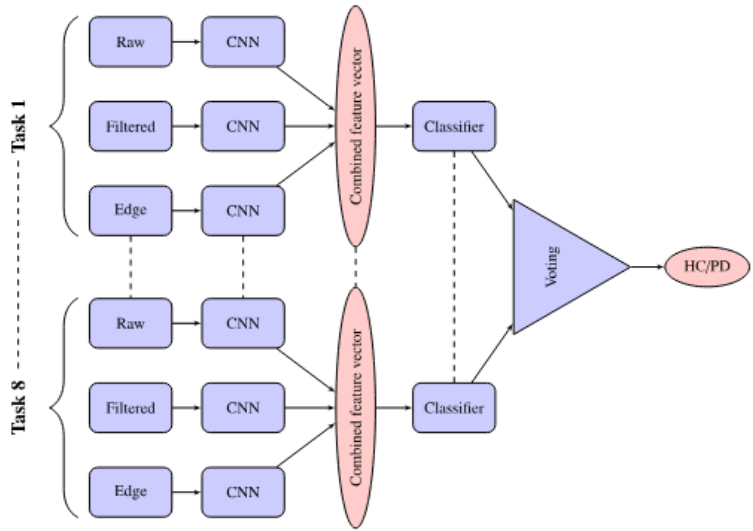

*Fig. 1. Classification workflow.*

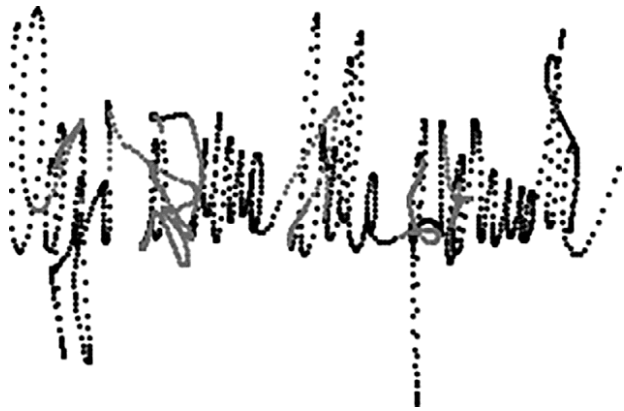

*Fig. 2. Enhanced image of a sentence written by a healthy young adult. The sentence is Oggi è una bella giornata ("Today is a beautiful day" in Italian).*

### 3.2. Feature extraction

An approach to extracting visual features from very small datasets is to use pretrained deep neural networks. The main rea- son is that, if the pretrained network has been previously trained on a large and general enough image dataset, the features learned by this network can be profitably used to identify and extract fea- tures from new images. In other words, the pre-trained network serves as a generic model of the visual world. This technique is an example of "transfer learning" and provides a robust alternative to more traditional approaches based on "hand-crafted" features.

In the present paper, the VGG16 architecture has been used [23]. This is a well-known and widely applied CNN architecture, trained on the very large ImageNet dataset. To achieve transfer learning, the densely connected layers on top of the network, which perform classification, were removed, and the convolutional base (up to *block5_pool*) was run for every input image of the PaHaW dataset. In the present study, in particular, the VGG16 model pre-packaged with Keras, was used [4].

Before feeding them into the CNN model, it was necessary to resize the images to $150 \times 150$ pixels, which is a common choice. In addition, since we used the same weights of the colored ImageNet dataset, each gray-scale image was converted to the RGB format.

Note that, as in [16], we enhanced feature learning by considering additional representations of the input data. Employing multiple representations helps increase the complexity of the data, thus enabling the learning of better-quality features. In particular, three representations were considered, each providing the input to one of three different convolutional networks:

- *Raw images*: these are artificially reconstructed images obtained as described in the previous subsection.

- *Median filter residual images*: these images were obtained by applying a median filter with a $3 \times 3$ window-size on the previous raw images and by subtracting the raw images from the corresponding filtered images.

- *Edge images*: the third network was trained on only the edge information of the raw images. The edges were obtained by applying linear convolution filters in the horizontal and vertical directions.

Each network constructs a hierarchy of visual features, starting from the earlier layers [4]. The first layers learn very simple concepts such as basic shapes. The internal representation of the deeper layers is difficult to interpret but consists of higher-level concepts constructed using the lower ones. The final output con- sisted in 8,192 features for each representation modality, so having a combined feature vector of 24,576 features for each task.

It is worth remarking that, since the set of data at our disposal is limited, we did not train the CNN models over these data, but we used them only as feature extractor.

*3.3. Model fitting*

The combined feature vectors extracted by the three CNN models, one for each task, were then fed into traditional machine learn- ing algorithms. Several studies employing this technique have re- ported successful results, e.g. [17]. Finally, the predictions obtained by the classification models trained on the different tasks were pooled together, via majority voting, to achieve the final classification. The models we used, as well as the ensemble of classifiers, are briefly described in the following. In this paper, the scikit-learn implementation [19] of these algorithms has been used.

*Support vector machines,* the main idea behind support vector machines (SVMs) is to find a separating hyperplane with the largest minimal distance from the closest data points of either class. A new example is predicted to belong to a class based on which side of the hyperplane it falls [26]. To learn nonlinear decision boundaries, the data points are mapped to a higher dimensional space via a kernel function, e.g. the radial basis function (RBF) kernel. In this study, we experimented with both the linear and the RBF kernel. The bias-variance trade-off of the algorithm is governed by the fine tuning of the penalty parameter $C$ and the kernel coefficient $\gamma$ in the case of RBF kernel [12]. In this paper, $C$ was set to 1 and $\gamma$ to $\frac{1}{n}$, where $n$ is the number of features. These values are commonly found in the literature.

*Random forest and extremely randomized trees,* these models belong to a family of algorithms which rely on the *bagging* method for building several "base learners", usually decision trees, and then combining their predictions to provide the final classification. Random forest (RF) repeatedly selects a bootstrap sample from the training set, chooses random subsets of features, then fits a decision tree to this sample [3]. Because of this randomness, the bias of the forest increases, but, due to averaging, its variance also decreases. In extremely randomized trees (ET), randomness is taken a step further, by also completely randomizing the cut-point choice while splitting each node of a tree [11]. This allows the variance to be reduced a little more, at the expense of a slight increase in bias. In this work, 500 decision trees were used as base learners for both RF and ET: this is a typical choice.

*AdaBoost* Contrary to bagging methods, AdaBoost (ADA) relies on *boosting*: a sequence of base learners is fitted to the entire dataset. Then additional copies of the classifier are fitted to the same data but where the weights of incorrectly classified examples are iteratively updated [9].

In other words, at each iteration, every base learner is forced to focus on the examples that were missed by the previous learners in the sequence. As base learners, we used 500 decision trees.

*Ensemble* Similar or conceptually different classifiers can be combined through a majority voting scheme, so that the individual weaknesses of each classifier are mitigated. Combining the predictions generated by different classifiers is likely to provide better predictions, due to diversification. Firstly, each of the above classification models was trained on every task individually and the performances obtained evaluated. This served to explore the most discriminating tasks between the eight originally proposed. We evaluated the ensemble obtained by combining the best five tasks, i.e. the ones achieving the highest prediction accuracy. Such task selection is likely to improve classification performance, as some tasks may be less useful for diagnosis than others and their presence may introduce additional bias in the data.

*Table 1: Single task performance: static handwriting. In bold the best accuracy per task, in italic the overall best five tasks.*

| Task | SVM lin. | SVM RFB | RF | ET | ADA |
|---|---|---|---|---|---|
| 1) Spiral | 59.58% | 60.00% | ***60.83%*** | 55.00% | 56.67% |
| 2) *l l l* | 36.25% | **57.91%** | 47.08% | 46.67% | 43.33% |
| 3) *le le le* | 56.67% | 61.67% | 63.33% | ***63.75%*** | 57.08% |
| 4) *les les les* | 48.33% | **54.16%** | 47.08% | 43.75% | 49.58% |
| 5) *lektorka* | 43.33% | ***59.58%*** | 52.91% | 52.08% | 47.50% |
| 6) *porovnat* | 51.25% | **58.33%** | 54.58% | 56.67% | 55.00% |
| 7) *nepopadnout* | 51.67% | 61.25% | 58.33% | 57.50% | ***65.00%*** |
| 8) Sentence | 58.33% | 59.16% | ***65.41%*** | 63.75% | 57.91% |

### 3.4. Model validation

The classification performance was validated through a 10-fold cross-validation. This practice is typically preferred with small datasets, as in our case. The set of examples was divided into ten folds: one fold was treated as test set; the remaining folds formed the training set. The splitting was *stratified* by diagnosis, so that each fold contained roughly the same number of subjects from each diagnostic group. The entire procedure was iterated 10 times, until each fold was used once as test set.

### 3.5. Feature selection

Since the number of features was disproportionately higher than the cardinality of the dataset, to reduce the dimensionality of the feature space of every task and to alleviate overfitting, a feature selection algorithm was applied before classification. The discriminating power of each feature was evaluated by considering its accuracy in separating PD from HC when used as a single input feature to a linear SVM classifier. All features were then ranked in accordance with this score and only the $n$ features providing the highest score were retained for the final model fitting.

It is worth remarking that feature selection was *nested* within cross-validation, so that the most discriminating features were chosen based only on the training set and were blind to the test set. Applying *a priori* supervised selection of features inadvertently introduces a bias in the classification model which may lead to overoptimistic results [12].

# 4. Experimental results

In the following subsections, we provide experimental results aimed at answering two main research questions:

*Is the proposed method more effective for the purpose of PD classification than the individual handwriting modalities (static/dynamic)?*
*Is the proposed method the best way to combine static and dy- namic features of handwriting for different tasks?*

## 4.1. Effectiveness against individual modalities

In the following subsections, the results of three experiments are reported:

1. The first experiment was concerned only with the use of static handwriting. It was conducted by using only on- surface trajectories during feature extraction.
2. The second experiment dealt with more classic hand-crafted dynamic features.
3. The third experiment used only dynamically enhanced static handwriting, i.e. the approach we propose in this paper.

### 4.1.1. Static handwriting

Table 1 reports the results obtained by each classification model on every task. The table, as well as the following tables, shows the mean accuracies in percentage terms, averaged over all the ten cross-validation iterations. These results concern only the use of static features of handwriting. Analogously to Moete- sum et al. [16], these features were obtained by using the same experimental setting described in the previous section, but with images reconstructed from only the on-surface trajectories of the handwriting. Since the sampling frequency of the device is high (200 Hz), the images obtained by linking these points can be considered similar to real static images.

*Table 2. Single task performance: dynamic handwriting. In bold the best accuracy per task, in italic the overall best five tasks.*

| Task | SVM lin. | SVM RFB | RF | ET | ADA |
|------|----------|---------|-----|-----|-----|
| 1) Spiral | 49.16% | **53.75%** | 45.83% | 49.58% | 46.67% |
| 2) *l l l* | 61.25% | 59.16% | 60.00% | 60.41% | ***67.08%*** |
| 3) *le le le* | 70.41% | 67.08% | 67.08% | 61.25% | ***72.50%*** |
| 4) *les les les* | 39.58% | **57.91%** | 48.75% | 45.83% | 53.33% |
| 5) *lektorka* | 49.16% | 52.91% | 54.58% | **60.41%** | 53.33% |
| 6) *porovnat* | 53.75% | 60.83% | 53.33% | 53.33% | ***63.75%*** |
| 7) *nepopadnout* | 53.33% | 53.33% | 55.41% | 49.16% | ***61.67%*** |
| 8) Sentence | 68.33% | 69.16% | 67.91% | **70.83%** | 67.91% |

*Table 3. Single task performance: dynamically enhanced static handwriting. In bold the best accuracy per task, in italic the overall best five tasks.*

| Task | SVM lin. | SVM RFB | RF | ET | ADA |
|------|----------|---------|-----|-----|-----|
| 1) Spiral | ***75.00%*** | 67.50% | 65.00% | 67.50% | 71.25% |
| 2) *l l l* | 58.33% | **64.16%** | 58.75% | 62.50% | 56.25% |
| 3) *le le le* | 53.75% | **58.33%** | 55.00% | 53.75% | 56.25% |
| 4) *les les les* | 57.08% | 61.67% | **71.67%** | 62.50% | 66.67% |
| 5) *lektorka* | **75.41%** | 69.58% | 74.58% | 75.00% | 70.41% |
| 6) *porovnat* | **63.75%** | 60.00% | 51.25% | 52.91% | 49.58% |
| 7) *nepopadnout* | 57.50% | 51.25% | 65.41% | 58.75% | ***70.00%*** |
| 8) Sentence | ***67.08%*** | 55.00% | 55.83% | 51.25% | 52.50% |

The five tasks obtaining the highest prediction accuracy were: drawing the spiral (task 1); writing *le le le* (task 3); writing the word *lektorka* and *nepopadnout* (task 5 and 7); and writing the entire sentence (task 8). The discriminating power of task 8 was expected and confirmed the findings reported in [7]: writing a long sentence probably requires more cognitive effort and so escalates the effect of the disease on the handwriting. Also the results concerning the repetition of the bigram *le* were expected: a hesitation between a character and the following one could point out the necessity to re-plan the writing activity, while fluid writing can reveal the presence of anticipated motor planning [2]. Conversely, the discriminating power of the spiral task contrasts with the findings of Drotár et al. [8], where the task was under- taken with no significant impact on classification. This may have been due to the use of measures only tailored to handwriting: the visual features provided by deep learning instead seem to overcome this issue. Similar observations were drawn in [16].

Table 4 reports the results obtained by the ensemble of the best five models over all tasks. In this table, we also report the mean values of the area under the ROC curve (AUC), the sensitivity and specificity. As expected, the ensemble improved on the individual accuracies obtained by the single tasks.

4.1.2. Dynamic handwriting

The results obtained by using dynamic features are reported in Table 2. It is worth noting that the same classification strategy, based on the ensemble of classifiers, was adopted. The main difference concerns the features used, which are the more classic, hand-crafted features previously proposed by Drotár et al. [7]. For reasons of space, their calculation is only sketched.

The horizontal and vertical components of the pen position were segmented into on-surface and in-air strokes, in accordance with the button status. Based on this segmentation, kinematic (displacement, velocity, acceleration, etc.), spatio-temporal (stroke size and duration, speed and stroke speed, in-air/on-surface time, etc.) and pressure-based features were computed. In addition, to uncover hidden complexities of the handwriting, features based on Shannon and Rényi entropy, signal-to-noise ratio and empirical mode decomposition were calculated. This feature extraction stage resulted in either a single value feature or a vector feature. For all the resulting vector features the following basic statistical measures were calculated: mean; median; standard deviation; 1st percentile; 99th percentile; 99th - 1st percentile (outlier robust range).

All features were normalized before classification, so as to have zero mean and unit variance.

As expected, an accuracy degradation in performing the spiral task was observed. Dynamic features tailored to handwriting seem not to be useful for a hand-drawing task. On the other hand, static and dynamic handwriting performance was consistent with the discriminating power of *le le le, nepopadnout* and the sentence task. The ensemble obtained with the best five tasks achieved better results than those obtained with static handwriting (see Table 4). This was expected, as dynamic handwriting is known to carry more information than static images of handwriting, so providing better performance.
This modality achieved a very high specificity (91.67%), which had a positive impact on AUC. A specificity higher than sensitivity indicates that the decision support system is better in detect- ing the absence of the disease in the healthy population rather than detecting the disease in the pathological group. This result confirms the findings we reported earlier [14], where dynamic features of handwriting were employed for the purposes of PD classification at earlier stages.

4.1.3. Dynamically enhanced static handwriting

Table 3 reports the results obtained by extracting features from the dynamically enhanced images. Compared to static and dynamic alone (Tables 1 and 2), the performance of the enhanced static handwriting showed an overall improvement. Surprisingly, the sentence task suffered an accuracy deterioration, while draw- ing the spiral and writing *lektorka* and *nepopadnout* achieved improved results.

It seems that the dynamic information injected into the static images helps discriminate in those cases where the task can be performed continuously without lifting the pen from the surface, as occurs in the spiral as well as in the single word tasks. This may have been due to the observation that healthy control handwriting is more fluent, while Parkinsonian handwriting is slower and segmented [2,25]. Therefore, in addition to slowness, it is characterized by in-air movement even when this movement is not required.

Conversely, in the sentence task, the in-air information is more prevalent than in the other tasks, at least because of the transitions from one word to the next. However, this prevalence may have hidden the underlying on-surface trajectory which may have been more discriminating in the previous experiment with dynamic handwriting.

The ensemble of the best five tasks outperformed static and dynamic handwriting alone in respect of accuracy (86.67%) and sensitivity (89.17%). It seems that the dynamically enhanced representation provides a very good sensitivity (higher than specificity). This is usually preferable in diagnosis systems, as a high sensitivity indicates that the system is able to rule out disease when resulting in a negative response.

In evaluating the individual contribution of the embedded dynamics, we note that including only temporal information (plotting the on-surface points instead of linking them) provided 81.25% of prediction accuracy, which is a performance comparable to that obtained with dynamic handwriting. This holds also for the general trend between sensitivity and specificity. This suggests that the improvement in terms of sensitivity, observed when combining the velocity and in-air information, mainly comes from the contribution of the in-air movements. The usefulness of in-air patterns in supporting the pathology evaluation has been exploited in previous works [1,22].

*Table 4. Ensemble of tasks performance. In bold the best results.*

| Handwriting features | Accuracy | AUC | Sensitivity | Specificity |
|---|---|---|---|---|
| Static | 72.50% | 65.83% | 78.33% | 64.17% |
| Dynamic | 81.67% | **87.29%** | 71.67% | **91.67%** |
| Dynamically enhanced static handwriting (*velocity*) | 81.25% | 86.32% | 75.83% | 86.67% |
| Dynamically enhanced static handwriting (*velocity* and *in-air*) | **86.67%** | 83.33% | **89.17%** | 80.83% |

*Table 5. Performance evaluation of different recombination methods. In bold the best results.*

| Recombination method | Accuracy | AUC | Sensitivity | Specificity |
|---|---|---|---|---|
| Template level, data augmentation, ensemble | **86.67%** | **83.33%** | **89.17%** | **80.83%** |
| Template level, no data augmentation, ensemble | 73.75% | 78.06% | 71.67% | 75.83% |
| Template level, data augmentation, fusion | 57.92% | 63.89% | 54.17% | 61.67% |
| Feature level, data augmentation, ensemble | 80.83% | 79.24% | 79.17% | 77.50% |
| Score level, data augmentation, ensemble | 72.92% | 74.50% | 71.00% | 75.67% |

*Table 6. Performance comparison with literature results on PaHaW.*

| Study | Handwriting features | Method | Accuracy |
|---|---|---|---|
| [7] | Dynamic | Single feature vector fed into SVM | 88.13% |
| [8] | Dynamic | Single feature vector fed into SVM | 81.30% |
| [16] | Static | Multi-expert system based on SVM | 83.00% |
| [18] | Dynamic | Single feature vector fed into RF | 72.39% |
| *This work* | Dynamically enhanced static | Multi-expert system based on different classifiers | 86.67% |

*4.2. Effectiveness against other recombination methods*

In the following, three additional experiments are reported:

Firstly, in order to evaluate the impact of the additional image representations (i.e., residual and edge images) as used in the classification performance, we applied the same method only to the raw images.

Secondly, we investigated whether the ensemble of the different tasks is beneficial rather than detrimental to prediction accuracy with respect to using a fusion of the features from all tasks. Finally, to evaluate if the proposed handwriting representation is the best way to combine dynamic to static handwriting, we experimented with combinations at both feature level and score level.

### 4.2.1. Effect of the additional representations

The method we propose provides additional representations of the raw images, namely residual and edge images, for the purposes of data augmentation. Previous work, e.g. Wang and Perez [27], emphasized the effectiveness of data augmentation on the final classification performance when using a deep learning approach. One problem with CNN models, in fact, is that they may have too few samples to learn from. As shown in Table 5, the additional representations greatly improved the predictive power of the proposed method.

### 4.2.2. Ensemble of tasks vs. fusion of tasks

An alternative approach to using an ensemble over the eight tasks is to fuse features first and then classify them later with a single classification model. In Table 5, we report the results of the latter approach based on the RF classifier. We report only the results obtained with RF, as the other classifiers used in this work provided slightly poorer performance. It can be noticed that the fusion of tasks dramatically reduces the overall performance with respect to the ensemble of tasks. This may be due to the very high dimensional vectors resulting from the fusion of all the eight tasks, which may have led to overfitting. However, the ensemble of tasks overcomes this issue. In addition, the ensemble approach may have allowed the model to mitigate the effects of tasks that are less relevant for the classification problem at hand.

### 4.2.3. Other combinations of static and dynamic

The dynamically enhanced static image representation here proposed represents a way to combine static and dynamic hand- writing at *reconstruction*, i.e. template, level. However, the two modalities can be combined in other ways, in particular:

-*Feature level*: for each task, static and dynamic features are combined into a high dimensional feature vector which is fed into a classifier pooled in the ensemble scheme.

-*Score level*: for each task, two classifiers are trained on static and dynamic features separately and their decisions are combined via soft voting. The predictions coming from the best tasks are finally merged via majority voting to achieve the final classification.

The results of this experiment are reported in Table 5. Concern- ing the combination at feature level, a performance degradation was observed. Maybe, as observed in the fusion of tasks, combining the high dimensional feature vectors coming from different sources of information provides very high dimensional feature spaces which cause the classification models to suffer from over- fitting. On the other hand, for what concerns the combination at score level, performance dramatically decreases, indicating that there is poor agreement between the analysis of static and dynamic handwriting. This can be explained by considering that the two handwriting modalities provide very different viewpoints of the same handwritten pattern. The dynamically enhanced static representation seems to be able to join them effectively.

*4.3. State-of-the-art comparison*

Table 6 summarizes the results obtained, comparing them with state-of-the-art results achieved on the PaHaW dataset. The best result is 88.13% [7] obtained with more traditional hand-crafted features. They used a simple approach based on feeding a single high dimensional feature vector, with features coming from all tasks (excluding spiral drawing), into a SVM-RBF classifier. This result was obtained with an *a priori* supervised selection of fea- tures applied before cross-validating the dataset. As described in Section 3.5, feature selection may or may not be "nested" within the cross-validation iterations. In the first case, the selection is carried out blindly to the test examples; in the second case, the procedure already sees the labels of the test examples and makes use of this information. This may lead to optimistic estimates of the true error rate. In our case, employing a "non-nested" feature selection within our method would result in an overoptimistic accuracy of 94.58%, along with a perfect AUC.

## 5. Conclusion

The handwriting of individuals suffering from Parkinson's disease is impaired. This suggests that handwriting analysis is a promising mean to developing intelligent systems for disease diagnosis and monitoring. When designing such systems, a crucial step concerns the choice of appropriate features to describe hand- writing. In this paper, the possibility of dynamically enhancing static images of handwriting has been investigated by embedding temporal and in-air movement information in the static images. The proposed combination provided improved results on the freely available PaHaW dataset.